\documentclass[journal,twoside]{IEEEtran}

\usepackage[table]{xcolor}
\usepackage{url}
\usepackage{listings}
\usepackage{graphicx}
\usepackage{enumitem}
\usepackage{amsmath}
\usepackage{amsthm}
\usepackage{amssymb}
\newtheorem{definition}{Definition}
\newtheorem{example}{Example}
\usepackage{amsfonts}
\usepackage[utf8]{inputenc}
\usepackage{multirow}
\usepackage[ruled,linesnumbered]{algorithm2e}
\usepackage{tikz}

\usetikzlibrary{calc,bayesnet,shapes,backgrounds,decorations.pathreplacing,shadows.blur}

\DeclareMathOperator*{\argmin}{arg\,min}

\begin{document}

\title{Towards automatic construction of multi-network models for heterogeneous multi-task learning}

\author{Unai Garciarena, Alexander Mendiburu, and Roberto Santana
\thanks{\textcopyright 2019 IEEE. Personal use of this material is permitted. Permission from IEEE must be obtained for all other uses, in any current or future media, including reprinting/republishing this material for advertising or promotional purposes, creating new collective works, for resale or redistribution to servers or lists, or reuse of any copyrighted component of this work in other works.}
\thanks{This work has received support form the predoctoral grant that Unai Garciarena holds (ref. PIF16/238) by the University of the Basque Country, and the IT-609-13 (Basque Government) and TIN2016-78365-R (Spanish Ministry of Economy, Industry and Competitiveness) programs http://www.mineco.gob.es/portal/site/mineco. 
We gratefully acknowledge the support of NVIDIA Corporation with the donation of a Titan X Pascal GPU used to accelerate the process of training the models used in this work.}
\thanks{Unai Garciarena and Roberto Santana are with the Department of Computer Science and Artificial Intelligence, University  of  the  Basque  Country  UPV/EHU, 20018 Donostia, Spain. 

Alexander Mendiburu is with the Department of Computer Architecture and Tecnology, University  of  the  Basque  Country  UPV/EHU, 20018 Donostia, Spain. 

e-mails: \{unai.garciarena, alexander.mendiburu, roberto.santana\}@ehu.eus}
}

\markboth{IEEE TRANSACTIONS ON NEURAL NETWORKS AND LEARNING SYSTEMS}{Garciarena \MakeLowercase{\textit{et al.}}: Towards automatic construction of multi-network models for heterogeneous multi-task learning}

\maketitle

\begin{abstract}
Multi-task learning, as it is understood nowadays, consists of using one single model to carry out several similar tasks. From classifying hand-written characters of different alphabets to figuring out how to play several Atari games using reinforcement learning, multi-task models have been able to widen their performance range across different tasks, although these tasks are usually of a similar nature. In this work, we attempt to widen this range even further, by including heterogeneous tasks in a single learning procedure. To do so, we firstly formally define a multi-network model, identifying the necessary components and characteristics to allow different adaptations of said model depending on the tasks it is required to fulfill. Secondly, employing the formal definition as a starting point, we develop an illustrative model example consisting of three different tasks (classification, regression and data sampling). The performance of this illustrative model is then analyzed, showing its capabilities. Motivated by the results of the analysis, we enumerate a set of open challenges and future research lines over which the full potential of the proposed model definition can be exploited.
\end{abstract}


\IEEEpeerreviewmaketitle

\section{Introduction}\label{sec:intro}

\IEEEPARstart{A}{rtificial} 
neural network (ANN) models  have seen their popularity rise and fall since they were proposed in the mid-20th century \cite{rosenblatt_perceptron:_1958}. These oscillations can be paralleled with the proposals of new and effective structures and training algorithms, and the results these propositions provided being surpassed by other models and methodologies. Now at its peak, their popularity has experienced its most recent increase thanks to the conception of deep neural networks (DNN), an ANN structure composed of several layers of neurons that interact with each other, and can have different structures and purposes inside the same model. These black-box models are theorized to perform deep learning (learning high-abstraction features of the data) that eventually produces high quality results in several domains. 

Due to their high modeling power, the form of DNN that we know today has broken through the barriers of the tasks they were designed for in the last millennium \cite{lecun_backpropagation_1989}; classification and regression of generic data structures with little or no exploitation of prior knowledge about said structures. Recently, DNN architectures were developed to manage different tasks with specific types of data, for example image classification using convolutional operations \cite{krizhevsky_imagenet_2012} or dealing with temporal data using recurrent connections \cite{hochreiter_long_1997}.

Despite the high modeling capacity of the DNNs, high-complexity tasks sometimes require more sophisticated model layouts. Because of this, multi-network models that can cope with these complex jobs were designed. One such paradigm is generative modeling, which consists of generating data similar but not equal to that known \cite{goodfellow_generative_2014,hinton_reducing_2006,kingma_auto-encoding_2013}. Even though single DNN models have been proposed for this type of task, the two most popular DNN-based generative models are two-network models; generative adversarial networks (GAN) \cite{goodfellow_generative_2014} and variational autoencoders (VAE) \cite{kingma_auto-encoding_2013}. These two models share similarities in terms of structural composition, as they both are composed of two individual - albeit connected - DNNs. The structural design of these two models was hand-made according to a predefined goal: the encoder-decoder structure in the VAE, and the generator-discriminator routine in the GAN. Even though extensions of these models have already been carried out \cite{ali-gombe_few-shot_2018,garciarena_expanding_2018}, these approaches were developed in restrictive frameworks that \textit{only} permitted relatively small, structure-wise modifications. These extensions consist of enhancing the model structure using additional networks that can serve different purposes to those already existing in the model.

In addition to multi-model proposals designed for high complexity tasks, other approaches attempt to comprise more than one functionality to a single network. This problem is known as multi-task learning (MTL) \cite{caruana_multitask_1997}, and consists of using one single model to learn several tasks of similar domains
\cite{liang_evolutionary_2018,meyerson_beyond_2017}, these domains usually consisting of classification, regression or reinforcement learning. The usage of DNNs to manage this kind of problems has produced interesting approaches in a very wide range of domains, e.g., using a single model to automatically play different Atari games with reinforcement learning \cite{fernando_pathnet:_2017} using \textit{supernetworks} or classifying characters from different alphabets \cite{liang_evolutionary_2018} using combinations of \textit{convolutional cells}.

In this paper, we address another, more general class of MTL, in which the different tasks not necessarily have to belong to the same domain (e.g., class prediction \textbf{and} data generation) and are solved at the same time. For this purpose, we take a step forward, introducing a model that can be composed of several interconnected DNNs of different types, capable of handling the heterogeneity in the set of tasks of the heterogeneous MTL (HMTL). We understand that such complex combinations of tasks require an advancement in the multi-network model design; a step forward in the automated generation of this kind of models. Therefore, we focus our efforts on providing a modeling scheme consisting of DNN building blocks placed in an interconnected structure, flexible and scalable, so that the model structure can easily be optimized; the VALP.

The main contribution of this work is threefold: (i) We provide a \textbf{formal definition of the VALP} as a general neural-network-based model to deal with HMTL. This high-abstraction definition aims at setting as few constraints as possible in terms of structural flexibility when designing a VALP implementation. (ii) From the abstract formalization, we present a \textbf{functional framework}  to illustrate the potential of the model. It is accompanied by \textbf{an example} of how a model instantiation that has to deal with three tasks of different characteristics can be created using the VALP definition. As it can be inferred, designing the structure of a model that can cope with such an extensive variety of problems is not a trivial task, and it is not examined in this work. Therefore, (iii) we thoroughly discuss several \textbf{open challenges and future research lines}.

The rest of the paper is organized as follows: We introduce the VALP definition in Section~\ref{sec:definition}. We then identify a set of components that can be included into any VALP implementation in Section~\ref{sec:instantiation}. In section \ref{sec:implementation}, we provide an illustrative example of a VALP instantiation. This is followed by the experimentation part that shows the viability of the VALP for the HMTL problem in Section~\ref{sec:experiments}. Regarding these results, we continue with a detailed future work part, in Section~\ref{sec:futurework}. Finally, conclusions drawn from this experimentation part can be found in Section~\ref{sec:conclusions}.

\section{VALP definition} \label{sec:definition}

\begin{definition}
A \textbf{data unit} is a pair $d = (\boldsymbol{v_d}, t_d)$. $\boldsymbol{v_d} = \langle v_d^0, v_d^1, ..., v_d^k \rangle$ represents the data with $k$ variables, which share the $t_d$ type.
\end{definition}

\begin{definition}
A \textbf{model input} is a data unit $i_j = (\boldsymbol{v_{i_j}}, t_{i_j})$ provided to the VALP.

$I$ is the set of all the inputs of the model: $I = \{i_0, i_1, ... i_p\}$.
\end{definition}

\begin{definition}
A \textbf{primary network} $n_w$ is a DNN. It is defined as a 5-tuple, $n_w=(i_{n_w}, f_w, a_w, p_w, o_{n_w})$. $i_{n_w} = \{ i^0_{n_w}, i^1_{n_w}, ... i^x_{n_w}\}$ is a set containing the inputs of the network, where each $i^j_{n_w}$ is a data unit. $f_w$ is a function representing how all $i^j_{n_w}$ are combined to form another data unit, the definitive input of $n_w$ (e.g., concatenation). The value of $i^j_{n_w}$ can vary over the different phases of the model life cycle (for example, a VAE decoder takes its input from an encoder during the training phase, and from a $\mathcal{N}=(0,I)$ when sampling). $a_w$ contains the type (e.g., Decoder) of the primary network, and $p_w$, its parametrization (e.g., hidden layer specification). $o_{n_w}$ is a data unit, and represents the output of $n_w$. It can be also considered an intra-model output.

$N=\{n_0, n_1, ..., n_y\}$ is the set containing all the primary networks in the model.
\end{definition}

\begin{definition}
A \textbf{model output} is a pair $o_j = (\psi_j, f_{o_j})$. $\psi_j = \{\psi_j^0, \psi_j^1, ..., \psi_j^d\}$ is the set of data units that $o_j$ receives from the networks in $N$. $f_{o_j}$ represents how all $\psi_j^i$ are combined to form the final $j$-th output of the model, a functionality similar to $f_w$.

$O$ is the set of all model outputs; $O = \{o_0, o_1, ..., o_r\}$.
\end{definition}

\begin{definition}
A \textbf{VALP} is a 4-tuple $M=(V, A, L, P)$. $V = I \cup N \cup O$, represents the model components. $A$ is a set of connections that determine how the model components are interconnected. $L = \{L_0, L_1, ..., L_q\}$ is a set of triples that defines how the model performance is assessed. $L_j = (l_j, p_{l_j}, \boldsymbol{g_j})$, where $l_j$ represents a loss function, $p_{l_j}\in \bigcup\limits_{n_w\in N}\{o_{n_w}\}\cup \bigcup\limits_{0<j<|O|}\{f_{o_j}(\psi_j)\}$ is a prediction (a data unit) made by the model (note that it can be either an intra-model or a model output), and $\boldsymbol{g_j}$ is the ground truth that $l_j$ uses to measure and improve the performance of the model with respect to a particular task.

$P$ represents the model hyperparametrization. It contains, at least, the parameters that specify how the different $L_j$ are combined to form a single loss function that can be used to optimize all the tasks of the model in a single step.
\end{definition}

\begin{definition}
A \textbf{model connection} is defined as $c_j=(i_{c_j}, o_{c_j}, \boldsymbol{\psi_{c_j}})$, where $i_{c_j}\in \bigcup\limits_{n_w\in N}\{o_{n_w}\}\cup I$ represents the data unit providing the information, and $o_{c_j}\in N\cup O$ represents the model component the information is delivered to. $\boldsymbol{\psi_{c_j}} = \langle \psi^0_{c_j}, \psi^1_{c_j}, ..., \psi^z_{c_j} \rangle$, $\psi^b_{c_j} \in  \mathbb{Z}$ $|$ $0\leq \psi^b_{c_j}<|v_{i_{c_j}}|$ represents indices of the variables transported from the connection input $i_{c_j}$ to the connection output $o_{c_j}$.

We define $A=\{c_0, c_1, ..., c_z\}$ as the set of all the connections of the model.
\end{definition}

The $V$ and $A$ sets of the $VALP$ can be used to form a directed graph (digraph) $G=(V,A)$. In a digraph, the number of arcs (connections) ending in a vertex is called the indegree of said node, whereas the number of arcs starting in a node is called the outdegree. Regarding these two characteristics, we differentiate three types of nodes in this digraph: (i) source nodes; those having an indegree value of 0, (ii) sink nodes; those with an outdegree value of 0, and (iii) internal nodes, which have both indegree and outdegree values strictly larger than 0.

In the VALP, the model connections are represented with arcs and there is a source node for each element in $I$, a sink node for each element in $O$, and an internal node for each element in $N$.

\section{VALP instantiation}\label{sec:instantiation}

Once the VALP has been formally defined, we identify a collection of elements that can be part of an instance of a VALP model. More specifically, we enumerate some key components indispensable for the correct operation of a VALP instance: data types and primary networks. We would like to make clear that the following lists neither intend to encompass all the items a VALP can be composed of, nor limit future additions to the pool of VALP components. In this regard, we expect the VALP to be able to embrace an extensive set of data types and a variety of primary networks.

\subsection{Data types}\label{sec:dataTypes}

The data type component in the data unit defined in the previous section presents an elegant manner to handle the heterogeneity required to the model in terms of data outputs. We therefore define four data types to which data units of a VALP could adhere to:

\begin{itemize}
    \item Discrete : This data type consists of a vector codifying discrete values.
    \item Numeric : This data type consists of a vector of numeric values.
    \item Samples: Similarly to Numeric, this data type consists of a vector of numeric values.
    \item Features: This data type also consists of a vector of numeric values.
\end{itemize}

Despite technically containing the same type of information, we choose to define Numeric, Samples, and Features as different types to improve the expresiveness of the model. For example, at the time of setting restrictions when creating a VALP instance (e.g., asserting that samples are provided where samples are due, and idem for other types, such as regression or classification), the data types will turn out useful, since, as well as improving the understandability of the operation of the model, they will improve the simplicity of said rules.

\subsection{Primary networks} \label{sec:nets}

Once we have identified a subset of data types that can be used within a VALP, we can similarly characterize a set of primary DNNs that can be included in the $N$ component of a VALP.

\begin{description}[labelindent=1cm]
	\item[Generic MLP, $g$:] A regular MLP that maps the provided input to an output. It can take any type of data unit as input. The data unit it produces can have different interpretations: numeric values (in any case) or samples (exclusively if it received samples). The activation function in the output layer is the identity function.

	\item[Discretizer, $\delta$:] Similar to a regular MLP, this network takes data units of any type as input, and produces data units of the Discrete values type. Its goal is to discretize values, mainly for classification purposes. It has a softmax activation function in the last layer.
    \item[Decoder, $d$:] The decoder receives Numeric or Feature data units, interprets them as means and variances of a $\mathcal{N}(\mu,I\times \sigma)$, and uses samples generated from that distribution to produce Samples data units. Its internal structure is also an MLP. 
    \item[Convolutional network, $c$:] This primary network exclusively consists of operations commonly found in Convolutional neural networks (CNNs): convolutional and pooling layers. It can only take and produce data of the type Feature. Its goal is to maximize the performance of a VALP instance when working with certain data structures  (e.g., image or sequential data).
    \item[Transposed-convolutional decoder, $t$:] This primary network can be seen as a combination of $c$ and $d$. It can only be composed of transposed-convolutional operations, and it can take Features, Numeric (guaranteed), or Discrete (optional) values. It produces Samples.
    
\end{description}

\subsection{Model loss function}\label{sec:lossinstance}

A straightforward approach for training the model is to use the regular backpropagation algorithm combined with a variant of the stochastic gradient descent (SGD) algorithm. These techniques optimize a loss function defined on the parameters of a model so that the performance of the model can be as close to perfection as possible. The VALP performs many approximations at a time, which means that various loss functions (with respect to both the model outputs and intra-model) need to be optimized in parallel. For a model that needs to optimize three different tasks (regression, classification, and data sampling), the following four kinds of loss functions have been identified as necessary.

\begin{itemize}
	\item Regarding the sample-generation outputs of a model ($t_{o^d_0}=$Samples), we need a loss function $l_0$ that can measure the likelihood of the model generating data that follows the distribution we are interested in.
	\item Related to the regression output of the model ($t_{o^d_1}=$Numeric), we need a metric $l_1$ that can compute the difference between two vectors of numeric values.
	\item For the classification outputs of a model ($t_{o^d_2}=$Discrete), we need a loss function $l_2$ that can compare discrete outputs with predictions of the same type.
	\item Regarding the output of the primary networks whose outputs are used in a $t$ or $d$, we need a metric $l_3$ that forces that output to approximate a distribution that we can reproduce.
\end{itemize}

It is important to note that some of these loss functions could have larger magnitudes than others, thus hoarding the effectiveness of the SGD algorithm. This could lead to some loss functions being ignored by the algorithm, producing a defective model. To address this issue, the VALP model includes one hyperparameter $\beta$, which scales the different sub-loss functions. This approach is inspired by the $\beta$-VAE \cite{burgess_understanding_2018}. It contemplates a $\beta$ parameter that scales the two sub-loss functions present in a common VAE.

\subsubsection{Data unit combination}
We also define example functions that can be used for the $f_n$ and $f_{o_i}$ presented in the VALP networks and outputs, respectively:

\begin{example}
Being, for example, $d^i=\{d^i_0, d^i_1\}$ a set of data units, we define the concatenation function as a function that receives a set of data units, and produces another one:

$\zeta(d^i, d^j) = (\langle v^0_{d^i}, v^1_{d^i}, ..., v^n_{d^i}, v^0_{d^j}, v^1_{d^j}, ..., v^m_{d^j} \rangle, t_{d^{i,j}})$

And, similarly, the addition function: 

$\Lambda(d^i, d^j) = (\langle v^0_{d^i} + v^0_{d^j}, v^1_{d^i} + v^1_{d^j}, ..., v^n_{d^i} + v^n_{d^j} \rangle, t_{d^{i,j}}), n=min(|v_{d^i}|, |v_{d^j}|)$.

In both cases, $(t_{d^{i,j}}=\text{Samples}) \leftrightarrow( (t_{d^{j}}=\text{Samples})\lor (t_{d^i}=\text{Samples}))$,

$(t_{d^{i,j}}=\text{Numeric}) \leftrightarrow ((t_{d^j}=\text{Numeric} \lor t_{d^i}=\text{Numeric})\land (t_{d^j}\neq\text{Samples} \land t_{d^i}\neq\text{Samples}))$,

$(t_{d^{i,j}}=\text{Discrete}) \leftrightarrow (t_{d^j}=\text{Discrete} \land t_{d^i}=\text{discrete})$.

$(t_{d^{i,j}}=\text{Features}) \leftrightarrow (t_{d^j}=\text{Features} \land t_{d^i}=\text{Features})$.

When $d^i = \{d^i_0\}$ consists of a single element, $f(d^i) = d^i_0, \forall f$.

\end{example}

\section{VALP instantiation}\label{sec:implementation}

The study performed in this work over the considers an implementation\footnote{The code developed to perform the initial exploration in this paper can be found in \url{https://github.com/unaigarciarena/VALP}} that consists of a reduced version of the general VALP defined in Section~\ref{sec:definition}. It does not cover all its possibilities, but rather is an initial exploration with the aim of displaying its potential, and has therefore many extension possibilities. The reduced VALP version considered in this work contemplates \textit{only} three of the data types introduced in the preceding section. We restrict all data types $t_d$ of all the data units in a VALP to the following values $t_d \in \{$Numeric, Discrete, Samples$\}$, $\forall d\in I \cup \bigcup\limits_{o_j\in O}\{o^d_j\} \cup \bigcup\limits_{n_w\in N} (i^{n_w} \cup \{o_{n_w}\})$.

Accordingly, only a subset of networks can be part of the VALP model definition of this work: $a_w\in \{g, d, \delta\}, \forall w\in \mathbb{Z}, 0\leq w\leq |N|$.

The Generic MLP essentially transforms information into a different \textit{encoding}, therefore, it can serve as an encoder that complements a Decoder. To avoid defining a primary network with the sole functionality of encoding data, we allow the interpretation of the output of any Generic MLP (a vector of numeric values) as the $\mu$ and $\sigma$ parameters a decoder needs. 

\begin{figure}
  \begin{center}
  \resizebox {8.8cm} {!}{\includegraphics{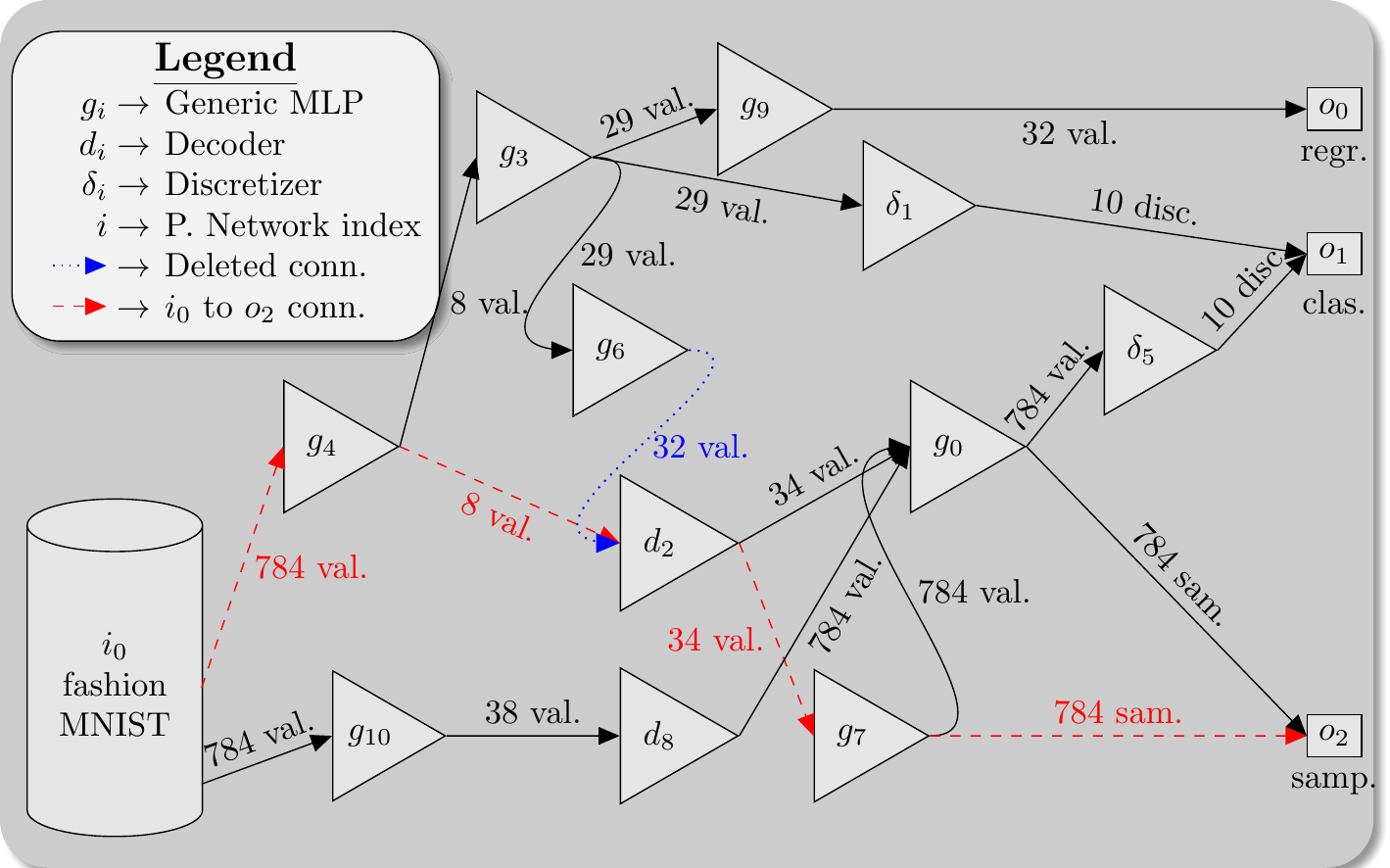}}
    \caption{Primary networks and their functionality inside VALP. ``Numeric" and ``Discrete" refer to numeric and discrete values. The decoder must take (at least) numeric values when training. When running the model, these values are replaced with samples from a $\mathcal{N}(0,I)$ distribution to ensure that the Decoder can create new data.}
    \label{fig:primNets}
  \end{center}
\end{figure}

Fig.~\ref{fig:primNets} describes these primary networks and the way they are related to the different data types.

Because we prime structural flexibility in the general VALP (and thus, in this example), the primary networks introduced in the previous section should be free to interact with each other in any possible way. However, in order to maintain type consistency (both in the output and throughout the model), a set of rules that restrict the model structure need to be imposed. In this regard, we force the decoders in a VALP configuration to have at least one Generic MLP primary network providing input to a decoder. This requirement is introduced due to the necessity of the decoders to have a numeric, optimizable input that can be trained to follow a certain distribution and can later be changed by new samples that follow said distribution.

\begin{figure}[ht]
\begin{center}
\resizebox {8.8cm} {!}{\includegraphics{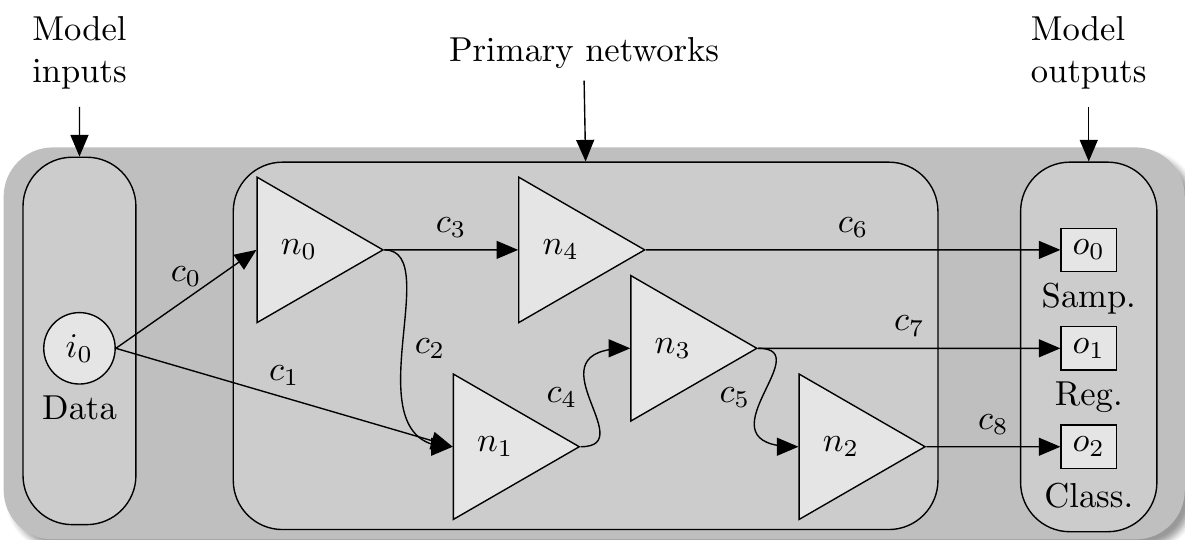}}
	
\caption{Schematic representation of a VALP.}
\label{fig:ModelExample}
\end{center}
\end{figure}

In Fig.~\ref{fig:ModelExample} an example of a VALP model designed for solving three different tasks (classification, regression and sampling) is shown. It is composed of five primary DNNs, it receives a single input $i_0$ (Data), and it provides three outputs, $o_0, o_1,$ and $o_2$, where $t_{o^d_0} = \text{Samples}$, $t_{o^d_1} = \text{Numeric}$, and $t_{o^d_2} = \text{Discrete}$. In the figure, circle nodes represent source nodes, triangular nodes represent internal nodes, and square nodes are sink nodes.

\subsection{Formal definition of a VALP instance}

In addition to the graphical illustration shown in Fig.~\ref{fig:ModelExample}, a formal definition of the example model is presented:

We define $I=\{i_0\}$. $i_0=(\boldsymbol{v_{i_0}}, t_{i_0})$, where $v_{i_0}$ are the 10 features of the database, and $t_{i_0}=\text{``Numeric"}$. 

We define $N=\{n_0, n_1, n_2, n_3, n_4\}$.

Because all the network types $a_w$ can take are based on MLPs, any $p_w$ is composed of three vectors: $init = \langle init_0, init_1, ..., init_l \rangle$, which specifies the function used to randomly initialize the primary network parameters, $act = \langle act_0, act_1, ..., act_l \rangle$, referring to the activation functions in each layer, and $ns = \langle ns_0, ns_1, ..., ns_l \rangle$, the number of neurons in each layer. $l$ is the number of layers in the primary network, excluding the input layer, which needs no parametrization, and including the output layer.

We do not define every aspect of each component in this VALP example for the sake of keeping the definition compact. We avoid definitions of elements that are repetitive or redundant and do not contribute to the further understanding of the model concept. For example, we do not define $p_w$ for each $n_w$.

$n_0 = (i_{n_0}, f_0, a_0, p_0, o_{n_0})$, where $a_0=\text{Generic MLP}$, 
$i_{n_0} = \{i^0_{n_0}\}$, $v_{i^0_{n_0}} = \langle v_{i^0_{n_0}}^0, v_{i^0_{n_0}}^2, v_{i^0_{n_0}}^5, v_{i^0_{n_0}}^6, v_{i^0_{n_0}}^7 \rangle$ (Note the correspondence later, when defining the connections), 
$t_{i^0_{n_0}} = \text{Numeric}$, and 
$f_0= \zeta$.
$o_{n_0} = (v_{o_{n_0}}, t_{n_0})$, where 
$v_{o_{n_0}} = \langle v_{o_{n_0}}^0, v_{o_{n_0}}^1, ..., v_{o_{n_0}}^6 \rangle $, 
$t_{o_{n_0}} = \text{Numeric}$.

$n_1 = (i_{n_1}, f_1, a_1, p_1, o_{n_1})$, where $a_1=\text{Generic MLP}$, $i_{n_1} = \{i^0_{n_1}, i^1_{n_1}\}$, $t_{i^0_{n_1}} = \text{Numeric}$, $t_{i^1_{n_1}} = \text{Numeric}$, and $f_1=\zeta$. $o_{n_1} = (v_{o_{n_1}}, t_{o_{n_1}})$, $t_{o_{n_1}} = \text{Numeric}$.

$n_2 = (i_{n_2}, f_2, a_2, p_2, o_{n_2})$, $a_2=\text{Discretizer}$,  $t_{o_{n_2}} = \text{Discrete}$, $f_2=\emptyset$

$n_3 = (i_{n_3}, f_3, a_3, p_3, o_{n_3})$, where $a_3=\text{Generic MLP}$, $f_3 = \zeta$

$n_4 = (i_{n_4}, f_4, a_4, p_4, o_{n_4})$, where $a_4=\text{Decoder}$, $t_{o_{n_4}} = \text{Samples}$. Note that, when training the model, $i_{n_4}=\{i^0_{n_4}=o_{n_0}\}$, but that changes at the time of running the model; $i_{n_4}=\{i^0_{n_4}=(x\sim\mathcal{N}(I,0), \text{Numeric})\}$

We define $O=\{o_0, o_1, o_2\}$. 

$o_0 = (\psi_0, f_{o_0})$, where $\psi_0=\{\psi^0_0=o_{n_4}\}$, and $f_{o_0}=\Lambda$

$o_1 = (\psi_1, f_{o_1})$, where $\psi_1=\{\psi^0_1=o_{n_3}\}$, and $f_{o_1}=\Lambda$

$o_2 = (\psi_2, f_{o_2})$, where $\psi_2=\{\psi^0_2=o_{n_2}\}$, and $f_{o_2}=\Lambda$

We define the model instance $M=(V, A, L, P)$ shown in Fig.~\ref{fig:ModelExample}. This VALP example is required to produce numeric and discrete predictions, as well as a sampling output. We assume that we are working with a single dataset ``Data", that is composed of 10 features, and where each example is labeled. The vector of these labels forms $\boldsymbol{C}$. Analogously, we have an $\boldsymbol{R}$ vector with a numeric value associated to each entry in the dataset. Finally, we have 5 extra features that we would like to reproduce (generate new samples), grouped in a vector $\boldsymbol{S}$ (one attached to each example in the Data, similarly to $\boldsymbol{R}$ and $\boldsymbol{C}$).

$L=\{L_0, L_1, L_2, L_3\}$, where $L_0=(l_0, p_{l_0}, \boldsymbol{g_0})$. $l_0$ is the log-likelihood function, $\boldsymbol{g_0}$ is the $\boldsymbol{S}$ data provided in the beginning of the problem definition, and $p_{l_0}=\zeta(\psi_0)=\psi_0^0 = o_{n_4}$. $L_1=(l_1, p_{l_1}, \boldsymbol{g_1})$. $l_1$ is the mean squared error (MSE) function, $g_1$ is the $\boldsymbol{R}$ data provided in the beginning of the problem definition, and $p_{l_1}=o_{n_3}$. $L_2=(l_2, p_{l_2}, \boldsymbol{g_2})$. $l_2$ is the cross entropy function, $g_2$ is the $\boldsymbol{C}$ data provided in the beginning of the problem definition, and $p_{l_2}=o_{n_2}$.  $L_3=(l_3, p_{l_3}, \boldsymbol{g_3})$. $l_3$ is the Kullback-Leibler divergence (KL), $g_3\sim \mathcal{N}(0,I)$, and $p_{l_3}=o_{n_0}$.

The hyperparameter of the model is a tuple of a single element $P=(\beta)$, which parametrizes $L$. $\beta$ is a set of tuples $\beta = \{(l_3, 0.5), (l_0, 0.8), (l_1, 0.9), (l_2, 1)\}$, where each tuple contains model components and a scalar. The loss functions defined in the model components and the scalar in the tuples are multiplied together, and then added up to form $0.5\times l_3 + 0.8\times l_0 + 0.9\times l_1 + l_2$, the definitive loss function used to train the model.

We define $A=\langle c_0, c_1, ..., c_8 \rangle$.

$c_0 = (i_{c_0}, o_{c_0}, s_{c_0})$, where $i_{c_0} = i_0$, $o_{c_0} = n_0$, $s_{c_0} = \langle 0, 2, 5, 6, 7 \rangle$ (note the correspondence between this connection and the previously defined $v_{i^0_{n_0}}$).

$c_1 = (i_{c_1}, o_{c_1}, s_{c_1})$, where $i_{c_1} = i_0$, $o_{c_1} = n_1$.
$c_2 = (i_{c_2}, o_{c_2}, s_{c_2})$, where $i_{c_2} = o_{n_0}$, $o_{c_2} = n_1$.
$c_3 = (i_{c_3}, o_{c_3}, s_{c_3})$, where $i_{c_3} = o_{n_0}$, $o_{c_3} = n_4$.
$c_4 = (i_{c_4}, o_{c_4}, s_{c_4})$, where $i_{c_4} = o_{n_1}$, $o_{c_4} = n_3$.
$c_5 = (i_{c_5}, o_{c_5}, s_{c_5})$, where $i_{c_5} = o_{n_3}$, $o_{c_5} = n_2$.
$c_6 = (i_{c_6}, o_{c_6}, s_{c_6})$, where $i_{c_6} = o_{n_4}$, $o_{c_6} = o_0$.
$c_7 = (i_{c_7}, o_{c_7}, s_{c_7})$, where $i_{c_7} = o_{n_3}$, $o_{c_7} = o_1$.
$c_8 = (i_{c_8}, o_{c_8}, s_{c_8})$, where $i_{c_8} = o_{n_2}$, $o_{c_8} = o_2$.

Although the introduced VALP model assumes fixed $I$ and $O$, we can think of a scenario where this is not the case. In this regard, the VALP is not a static model structure-wise, as $I$ and $O$ can be expanded once the model has been learned, which would trigger additions in either $A$ or $N$, or both. For example, if we added a new $o_j$ to $O$, we could update the model by adding a new $n_w$ to $N$ and two connections, one from $n_j, j\neq w$ to $n_w$ and another one from $n_w$ to $o_j$. Such a scenario could be particularly useful for incrementally learning VALP instances.

\section{Testing the potential of a VALP}\label{sec:experiments}

We designed an artificial problem to illustrate the potential of the VALP. We have selected the widely known Fashion-MNIST \cite{xiao_fashion-mnist:_2017}, which is more complex than MNIST and has been extensively used for research on DNNs \cite{ciregan_multi-column_2012,jarrett_what_2009}. This dataset consists of 60,000 train and 10,000 test images, which are 28 $\times$ 28 pixel, gray-scale images of clothing, each belonging to a certain class. There are 10 items of clothing overall; T-shirt, Trouser, Pullover, Dress, Coat, Sandal, Shirt, Sneaker, Bag, and Ankle boot. The usual supervised task associated to this dataset is to predict the class each example belongs to.

To illustrate the model instantiation at its fullest potential, we define the multitask fashion-MNIST problem as composed of three different tasks:

\begin{enumerate}
	\item Classification: 10-class classification as usually defined in the fashion-MNIST problem.
    \item Regression: Firstly, we have computed a histogram for each of the images regarding the gray-scale values, with 32 bins. Then, these histograms were scaled between 0 and 1. This way, we have a 32-value regression problem to solve.
    \item Generation: The generation of samples similar to those given to the model in the input.
\end{enumerate}

\begin{figure}
\begin{center}
\includegraphics[width=8.8cm,trim={2pt 1pt 2pt 4pt},clip]{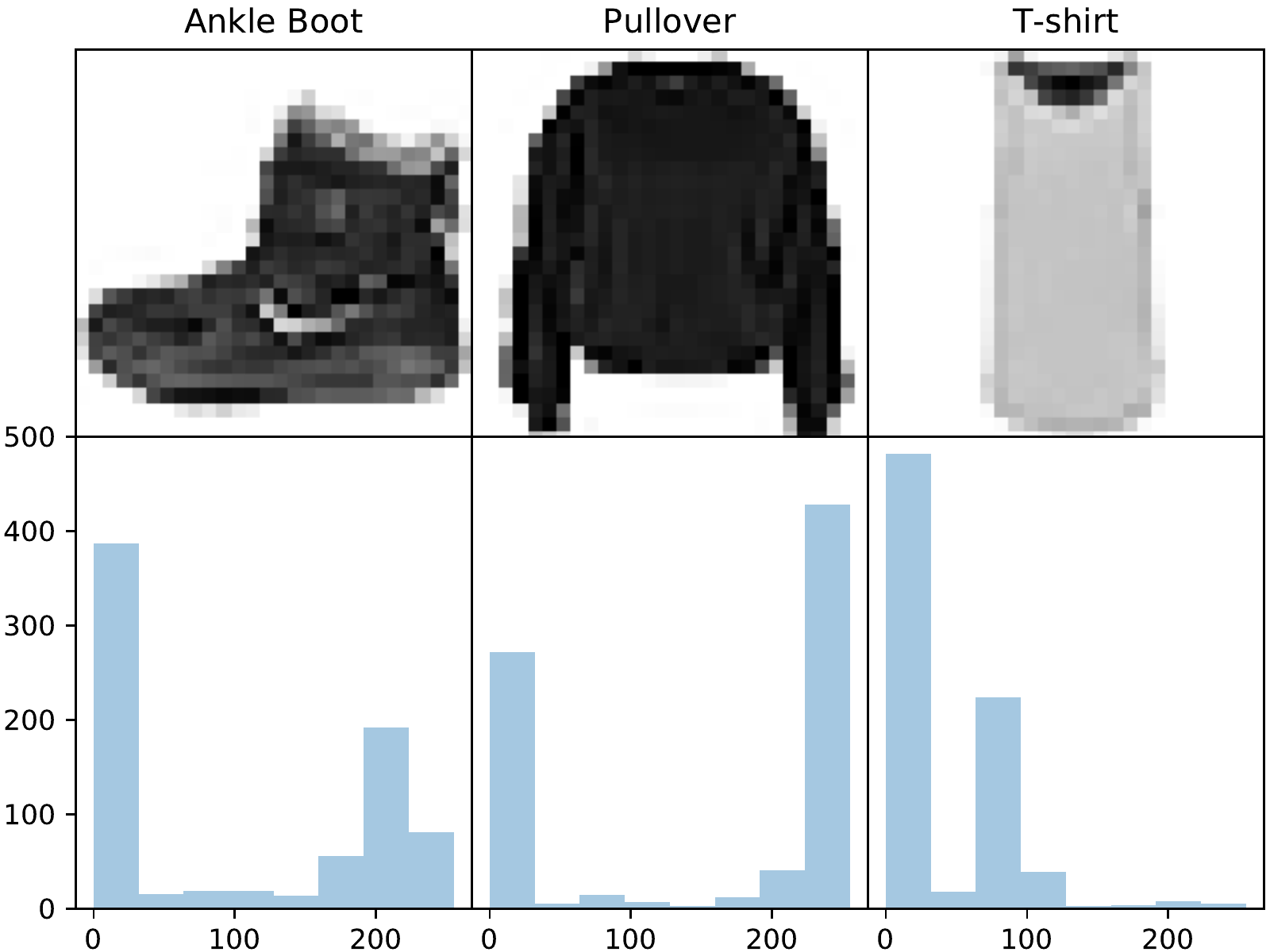}
\caption{This figure displays three gray-scale images that correspond to three different examples in the fashion-MNIST database. Their respective titles show what class they belong to (those used for the classification task), whereas the bar plots show the 8-bin-histograms representing the frequency of pixel values in the image (used for the regression task). The clothing image itself is used for the sampling task}
\label{fig:taskExamples}
\end{center}
\end{figure}

In Fig.~\ref{fig:taskExamples}, we show three examples of the data available in the fashion-MNIST dataset, with the values desired to be obtained for each one, in each task.

\subsection{Model parameters} \label{sec:modelParameters}

To initialize the model structure, the only information required is the number and types of inputs and outputs together with their corresponding dimensions. In our case, we can define $I=\{i_0\}$ and $O=\{o_0, o_1, o_2\}$, where $i_0=(v_{i_0}, t_{i_o})$, $|v_{i_0}| = 28\times 28=784$ and $t_{i_0} = \text{Numeric}$. $o_0=(\psi_0, f_{o_0})$ where $|\psi_0^0|=32$, and $t_{\psi_0^0}=\text{Numeric}$, $o_1=(\psi_1, f_{o_1})$ where $|\psi_1^0|=10$, and $t_{\psi_1^0}=\text{Discrete}$, and $o_2=(\psi_2, f_{o_2})$, where $|\psi_2^0|=28\times 28=784$ and $t_{\psi_2^0}=\text{Samples}$. $f_{o_0}=f_{o_1}=f_{o_2}=\Lambda$.

The next concern is to design a model (i.e., the primary networks and connections in the model) that can provide predictions for all required outputs, while matching the appropriate data type (Section~\ref{sec:dataTypes}).

The choice of the loss functions is a relevant decision to be made in this framework. As mentioned in Section~\ref{sec:lossinstance}, four necessary loss function types have been identified:

\begin{itemize}

    \item For the regression output, $L_0=(l_0, p_{l_0}, \boldsymbol{g_0})$, where $l_0$ is the mean squared error between $p_{l_0}$ and $\boldsymbol{g_0}$:
    	\begin{align}
    		\argmin_{\theta^{VALP}} \frac{1}{|p_{l_0}|}\sum(p_{l_0} - \boldsymbol{g_0})^2
     		\label{eq:LossReg}
		\end{align}

	\item For the classification output, $L_1=(l_1, p_{l_1}, \boldsymbol{g_1})$, where $l_1$ is the cross entropy between $p_{l_1}$ and $\boldsymbol{g_1}$:
    	\begin{align}
    		\argmin_{\theta^{VALP}} -\sum_x p_{l_1}\, \log \boldsymbol{g_1}
     		\label{eq:LossClass}
		\end{align}
        
	\item For the sampling output, $L_2=(l_2, p_{l_2}, \boldsymbol{g_2})$, where $l_2$ is the log-likelihood of $p_{l_2}$ being $\boldsymbol{g_2}$:  
    	\begin{align}
    		\argmin_{\theta^{VALP}} \mathbb{E}_{{\bf{x}} \sim \boldsymbol{g_2}} [\mathbb{E}_{q_\theta({\bf{z}}|{\bf{x}})} [-log(p_{l_2})]]
     		\label{eq:LossDec}
		\end{align}
		where $q_\theta({\bf{z}}|{\bf{x}})$ represents the probability distribution (predicted by a Generic MLP) inside the model, before any Decoder in a VALP.

	\item For each output of the generic MLPs whose output ($p_{l_3}$) feeds a decoder, $L_3=(l_3, p_{l_3}, \boldsymbol{g_3})$, where $l_3$ is the KL: 
    	\begin{align}
    		\argmin_{\theta^{VALP}} \mathbb{E}_{{\bf{x}} \sim p_{data}({\bf{x}})} [KL(p_{l_3}||\boldsymbol{g_3})] \label{eq:lossEnc}
		\end{align}
		As commonly, $\boldsymbol{g_3}\sim\mathcal{N}(0,I)$.
\end{itemize}

This model also considers different optimization pressures \cite{burgess_understanding_2018} on each of the terms of the global loss function. We add a scaling vector parameter ($\beta$), so that the optimization of the combined loss function is correctly performed.

\subsection{VALP structure designing algorithm}\label{sec:config}

For this example, we have designed a procedure for creating the VALP structure. As it will be discussed in Section~\ref{sec:futurework}, the conception of efficient algorithms for designing VALP structures is an interesting open challenge. 

The structure initialization algorithm employed in this work, which follows a back-to-front building approach, ensures that the adequate data type is provided to each model output. The strategy is based on an updated set of \textit{model components} that have not had an input assigned; $act\_cmp\subset N\cup O$. The algorithm is a recursive function that incrementally and randomly develops the model until the maximum number of primary networks for the model (a parameter given at initialization) is met. It produces a configuration in which not having a component without an input for it is guaranteed, and the output types match the requirements. The model outputs can receive their predictions from a network that produces the data type that satisfies it, while the networks can receive inputs from either another network, or a model input, i.e., the data. The only restriction applied to this algorithm is that a decoder must have at least one Generic MLP providing input. The reason behind this constraint is that the input of a decoder must be numeric and optimizable, as we require it to follow a certain continuous distribution ($\mathcal{N}(0,I)$). 

The pseudo-code of this strategy is shown in Algorithm \ref{alg:initialization}.

\begin{algorithm}[ht]{
	
	{\fontsize{10.08pt}{12.016}
	\SetKwFunction{FMain}{initialize}
  \SetKwProg{Fn}{Function}{:}{}
  \Fn{\FMain{$A$, $I$, $N$, $O$, act\_cmp}}{
        \If{($\text{max\_n}-|N|$) == $|\text{act\_cmp}\cap O|$\label{alg:if1}}{\Return complete\_model(model)}
        con\_out = random($O\cup N$)\\
        found, con\_in = random($I\cup N$, con\_out)\\
		\If{random\_numb(0, 1)$<\alpha \lor \neg\text{found}$\label{alg:if2}}{
        con\_in = create\_rand\_network(c)\\
        $N=N\cup \{\text{con\_in}\}$\\
        act\_out $= \text{act\_out} \cup \{\text{con\_in}\}$\\
        }
        $\psi = $random\_choice($con\_in$)\\
        $A = A \cup {(\text{con\_in}, \text{con\_out}, \psi)}$\\
        $\text{act\_out} = \text{act\_out} - \{\text{con\_out}\}$\\
        
        \Return initialize($A$, $I$, $N$, $O$, act\_cmp)
  }
	}
}
	
	\caption{Model structure initialization algorithm.}
	\label{alg:initialization}
\end{algorithm}

This algorithm considers two parameters; max\_n$=11$, which handles the maximum number of primary networks in a VALP, and $\alpha=0.5$, which regulates the reutilization of primary networks. More specifically, the $\alpha$ parameter is used to decide whether a source of data $c_1$ (if available) is used as an input for another component $c_2$ ($c_1\in I\cup N$, $c_2\in N\cup O$).
In addition to these parameters, this algorithm uses a set of auxiliary functions, which are explained below:

\begin{itemize}
    \item complete\_model(model): If $|N|$==max\_n, then act\_cmp==$0$, and the function does nothing. If $|N|<\text{max\_n}$, it searches for components that can serve the elements in act\_cmp (in terms of data typing) and establishes connections between them. If no such components exist, this function creates new primary networks, and uses them as bridges between the available components and those in act\_cmp.
    \item random\_numb(a, b): It returns a random number in $[a,b)$.
    \item random(set[, out]): If only the set parameter is provided, this function returns a random element from the set. If both parameters are present, it returns a random component from the set, such that it can serve as input to the out component (in typing terms, and not allowing recurrent connections) and found=$True$. If no such element exists, found=$False$.
    \item random\_choice(i): This function draws a random amount of numbers $n$ from $0\leq n<j$, without replacement. $j = |v_{con\_in}| \leftrightarrow i\in I, j = |v_{o_{n_w}}| \Leftrightarrow i\in N$
\end{itemize}

The initial call to the recursive algorithm is \texttt{initialize}$(\{\}, I, O, \{\}, act\_cmp)$, where act\_cmp is a copy of $O$.

The first \textit{if} statement (line \ref{alg:if1}) is the exit condition. In case it is not met, the algorithm selects a random component ($con\_out$) that can take an input, i.e., a model output or a primary network. $con\_out$ will have a new input once the actual recursion is finished. The algorithm searches for another component ($con\_in$) that can provide a data unit which can serve as an input for the first one. 

If no such element is found, or if a random number is lower than the $alpha$ parameter (line \ref{alg:if2}), a new primary network ($con\_in$) that can serve as input to the first component is created and added to $N$ and act\_out. Finally, $con\_out$ and $con\_in$ are connected. Because $con\_out$ now has an input, it cannot be part of act\_out. This algorithm avoids recurrent connections.

For example, when using this algorithm to create the example displayed in Fig.~\ref{fig:ModelExample}, the first step would consist of $N=\{\}$ and act\_cmp=$\{o_0, o_1, o_2\}$. In the first recursion, $n_3$ could be added, a generic MLP, since it needs to provide a regression prediction. In the following recursion, act\_cmp=$\{o_0, n_3, o_2\}$, because $o_1$ already has $n_3$ giving it an input, and $n_3$ would be added. This recursion would end up with at least three primary networks providing information to the outputs, and no component without a data input, act\_cmp$=\{\}$.

Once the model structure is defined, the parameters of the primary networks (weights and biases) are randomly initialized and trained with regular backpropagation, taking as the global loss function a combination of the elements in $L$ (according to the $\beta$ parameter).

\subsection{Experimental design}

For the multitask fashion-MNIST problem, we performed a preliminary experiment and found that there were no problems with optimizing $L_0, L_1, L_2$ with $\beta_0, \beta_1, \beta_2=1$. $L_3$, however, did present an obstacle with optimizing $L_0, L_1,\text{ and } L_2$. Therefore, we set $\beta_3=10^{-4}$. This way, we prevent the oscillating effect of the KL loss function rendering the optimizer futile once $L_0, L_1, L_2$ reach a value below that of $L_3$.

The max\_n was arbitrarily set to 11. The mini-batch size for training the model was 50, and the model was trained for 40,000 epochs. Therefore, $P=(\beta=(\beta_0=1, \beta_1=1, \beta_2=1, \beta_3=10^{-4}), \alpha=0.5, \text{max\_n}=11)$.

In order to have a glimpse of the performance of this VALP instantiation, the structural generation and learning procedures were randomly initialized and performed 500 times. This way, we perform a random search over 500 generable structures.

\subsection{Experiments results}

After training the 500 random VALP configurations, we employed each model to perform the described tasks considering the test set of the fashion-MNIST dataset, and recorded different metric values to get a measurement of their performance.

\subsubsection{Classification and regression}

\begin{figure}
\begin{center}
\includegraphics[width=8.8cm,trim={2pt 3pt 2pt 0pt},clip]{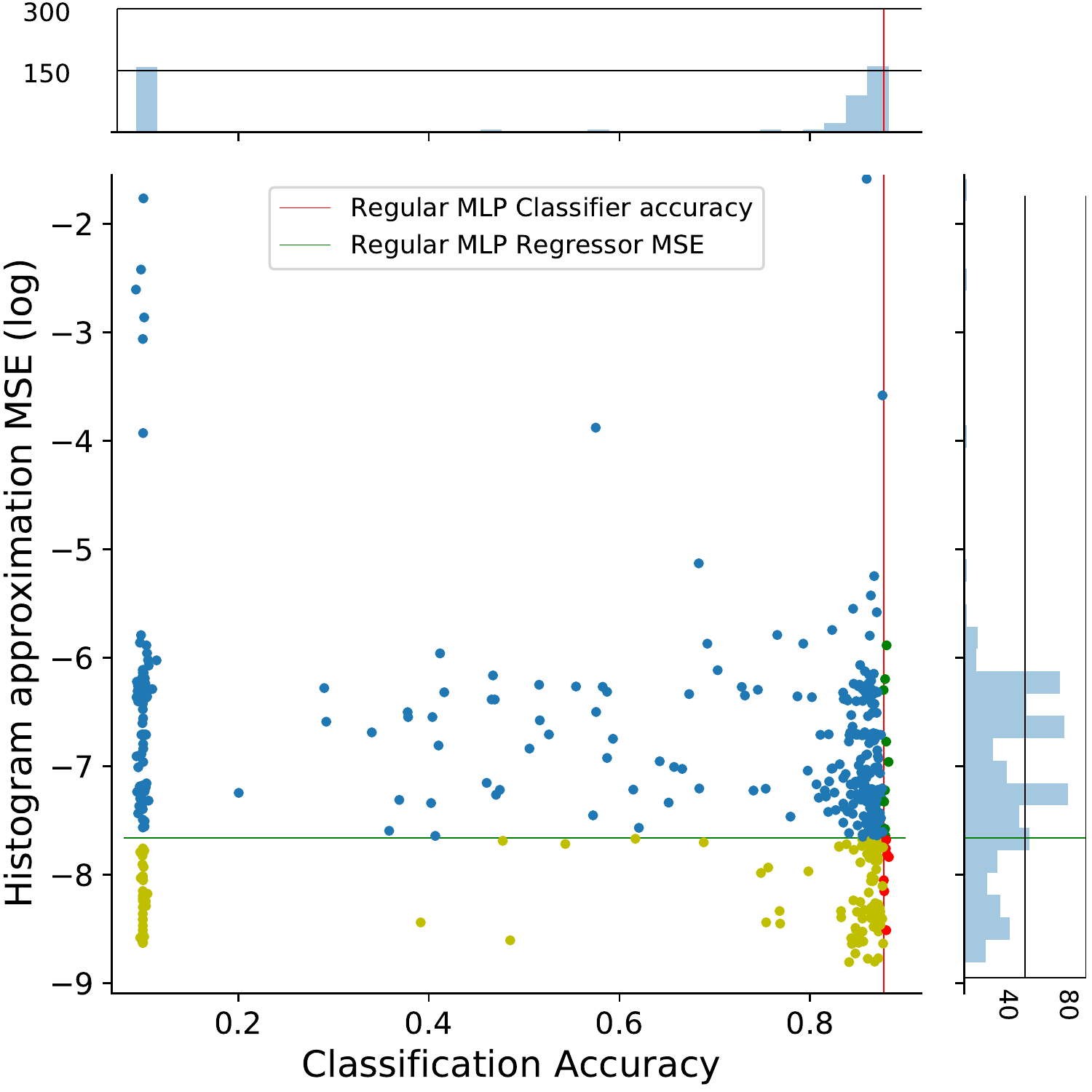}
\caption{Model performances plotted in terms of both classification accuracy and MSE reached on the regression problem. The vertical red line represents the performance of the MLP specifically learned for the classification problem. The horizontal green line represents the performance of the MLP specifically learned for the regression problem. Blue points represent VALP configurations that performed worse than both baselines, yellow points are models that improved only the regression problem results, and models represented in green did so in the classification problem, and red ones offered better results than both baselines.}
\label{fig:mseacc}
\end{center}
\end{figure}

In Fig.~\ref{fig:mseacc}, we can observe how the 500 models have performed on the regression (MSE represented on the y axis in a logaritmic scale) and the classification problem (accuracy, on the x axis). Each point in the grid represents a random VALP configuration. 

The horizontal green line represents the performance of a 100-neuron single-hidden-layer MLP with ReLU activation function trained exclusively for the regression problem. Analogously, the vertical red line represents the classification accuracy of the same model (with a softmax activation function in the output layer). Note that the goal of this experimental section is not, by any means, the comparison of the VALP against other models devoted to specific tasks. These are displayed solely to give an idea of where the modeling power of this particular VALP specification stands.

Regarding the classification problem, we can see that 158 VALP configurations produced a classification accuracy of nearly $10\%$. Considering that there are 10 classes, this resembles random classification. The architectures producing these results probably have a decoder that deletes the path between $i_0$ and $o_1$ after training. The other weak outcomes are probably a result of similar structures, or models whose optimizing algorithm has focused on the other loss functions involved. As an additional observation, we can see that the classification prediction is competitive with the baseline: 215 instances above $85\%$. It has to be taken into account that the VALP is managing at least four different loss functions at the same time, whereas the MLP is focusing only on classification.

With respect to the regression problem, we can observe that only 9 configurations performed poorly compared to the vast majority of the models; MSE superior to $0.007$. Contrasting these results with the baseline MLP, which had a $0.00047$ error, we can observe that $131$ models produced better results.

As a general remark on the joint performance, we can observe that 7 models performed better than both baselines (in classification and regression). The fact that a random search was able to obtain VALP configurations that have stronger performances in both objectives, while optimizing the sampling loss functions at the same time, shows the validity of the proposal at tackling HMTL problems.

\subsubsection{Generation}

Once it has been shown that the VALP can perform the tasks classically attached to the \textit{simple} DNNs, it is time to investigate the sampling capabilities of the model. Specifically, we want to extract information of two different aspects from the generative power of the model, which are related to the mode collapsing problem. Mode collapsing is an issue that concerns the generative modeling community, specially that part focused on GANs \cite{srivastava_veegan:_2017}. In the experimentation carried out in this paper, we identify two types of mode collapsing. The \textit{global} case, the worst one, is that in which all the generated samples look very similar to each other. In the \textit{local} mode collapsing scenario, not as serious as the previous one, the model would learn to generate samples from the different classes of the dataset, yet those belonging to a certain class are still too similar to each other.

To test whether the model configurations suffered from global mode collapsing, an auxiliary DNN that classifies samples of the fashion-MNIST was trained. This DNN was used to classify the generated samples, giving a metric of how distributed the generations are class-wise. This DNN was based on the MobileNet model \cite{howard_mobilenets:_2017,jianlin_baseline_2017}, followed by a single dense layer. This model reached a 99.5$\%$ accuracy on the training set and 94.5$\%$ on the test set and it was used to classify the samples generated by each VALP configuration. In terms of classes, we consider a sampling model to be perfect if it is able to obtain the same class distribution it has been shown. Given that the classification problem being addressed is balanced, one could expect that the perfect model generated the same number of instances of each class. Because this is a 10-class problem, a perfect generator would generate an example of a certain class with $0.1$ probability. Once the labels from the classifier were obtained, the capacity of the model of generating samples from different classes was measured as the entropy of the set of predicted labels. The higher the entropy value which ranges between 0 and 1, the better the model is considered.

\begin{figure}
\begin{center}
\includegraphics[width=8.8cm,trim={3pt 1pt 2pt 5pt},clip]{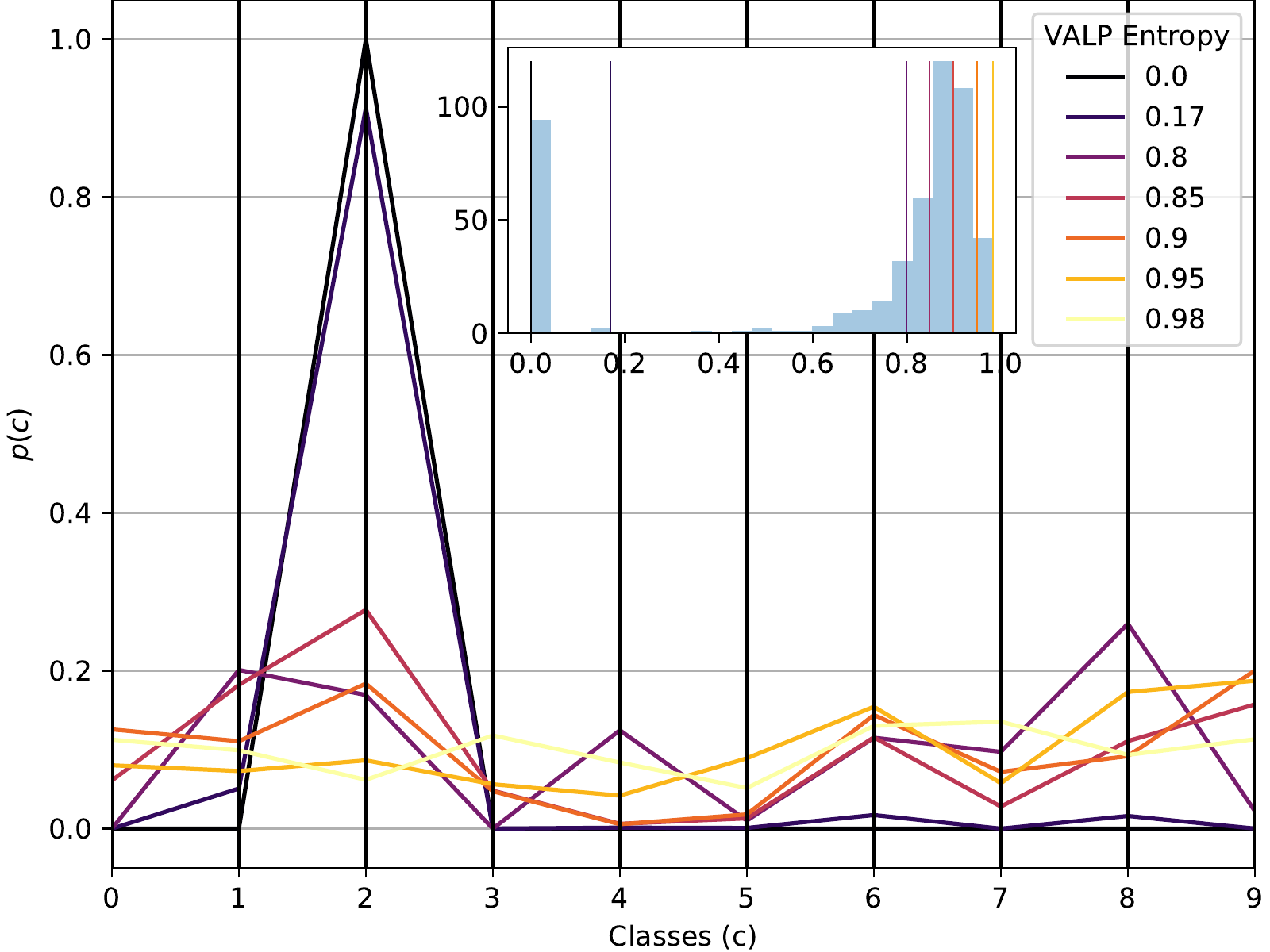}
\caption{The box in the top shows the distribution of entropy values and seven selected VALP examples. Each representative example is displayed by a vertical line. The probabilities of generating a sample of a certain class by each representative example are shown in the general parallel coordinates.}
\label{fig:parallel}
\end{center}
\end{figure}

The class distribution can be visualized in Fig.~\ref{fig:parallel}. The figure in the top represents the distribution of the entropy values of the 500 VALP configurations. We have chosen 7 representative examples from the whole set (pictured as vertical lines). The probability distributions of said examples are represented in the general parallel coordinates. In this example, we can observe how several VALP configurations (low entropy values) are poor data samplers, as they tend to generate images that MobileNet classifies as Pullover (class 2). This category usually consists of a set of pixels with high values in the middle of the image. The usual result of poor generators is producing images that consist of means of all the images which results in blurry images that posteriorly MobileNet classifies as Pullover. A similar effect happens with class 6 (Shirt). In contrast, the VALP configurations with higher entropy values (equal or larger to $0.8$) show much more distributed generation probabilities.

\subsubsection{Conditioned sampling}

Regarding the sampling capabilities of the VALP, we recognize that some problem domains could benefit by being able to \textit{force} the model to create data that meets certain criteria. This is, rather than creating it from the whole known space, data is created from a specific sub-region of the whole space. By designing the VALP as highly flexible as it is, we have made the VALP instances in this work capable of carrying out this specific kind of sampling. This is an interesting characteristic that can be exploited for various tasks, such as generating images with certain attributes \cite{yan_attribute2image:_2016} or reconstructing corrupted images \cite{isola_image--image_2017}.

To address this issue, we added a parameter $\phi$ that regulates the probability of a decoder not having an input deleted, i.e., we can have conditioned decoders that receive information from the model inputs, even when the training phase has finished. This parameter can only affect the model when a decoder receives more than one input, and it guarantees that at least one input is deleted.

To test how conditioned the samples can be towards the input of data given to the VALP, we computed a precision metric that evaluates the agreement (measured as an accuracy) between MobileNet's prediction for the test image used to condition the generation and the prediction for the corresponding VALP generated sample.

\begin{figure}
\begin{center}
\includegraphics[width=8.8cm,trim={0pt 2pt 5pt 0pt},clip]{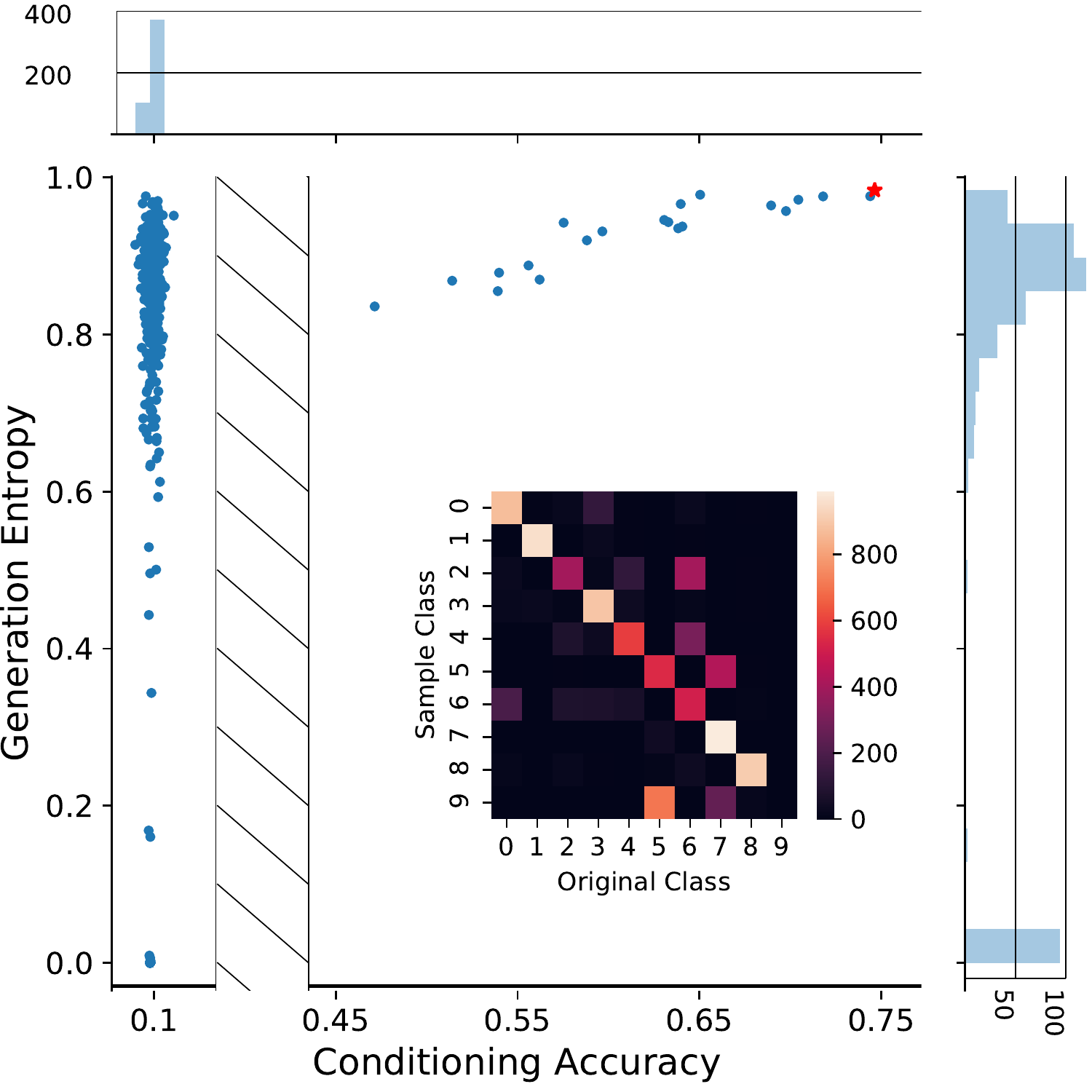}
\caption{Sampling performance of the 500 VALPs according to conditioning capability and distribution over the possible classes. The red star represents the VALP configuration with the largest conditioning accuracy. It also displays a heatmap representing a confusion matrix, comparing the class of the conditioning examples and the generated sample images by the VALP represented with the red star.}
\label{fig:entracc}
\end{center}
\end{figure}

The results of this metric along with the entropy can be observed in Fig.~\ref{fig:entracc}. Each VALP configuration is represented by a point, and they are located in the grid regarding the class entropy and the conditioning accuracy. It can be observed that most models have a high entropy value, which means that there is not a strong global mode collapsing problem. Part of the figure has been cut out, as there were no models that generated \textit{conditioning values} between $0.15$ and $0.45$. Additionally, this figure shows a heatmap representing the confusion matrix comparing the class that MobileNet predicted to the conditioning examples, and the class the Mobilenet predicted for the produced samples by the VALP represented with the red star.

Firstly, we observe that \textit{only} about $100$ models generated poor class entropy, therefore, in that aspect, the models created offered a good behavior. Fig.~\ref{fig:entracc} also shows that there aren't many models that can generate conditioned samples. This was to be expected, as there is not much probability of producing a configuration with a decoder that has more than one input, and from those configurations, only an expected $30\%$ would implement conditioning structures. Overall, it is easy to distinguish which models have a sampling conditioning component, as there is a big gap from $~10\%$ (where the unconditioned models are found) to $45\%-75\%$, (where the conditioned models are shown). Overall, the random initialization algorithm, was able to create $21$ models capable of generating conditioned samples, those in the right-hand side of the figure. The rest, those near $0.1$ \textit{conditioning accuracy}, are unconditioned samplers. The configurations in the lower part of the left-hand side of the figure are the poor quality generators. The model with the highest conditioning accuracy ($\sim0.75$) also happened to be the one with the largest generation entropy, and it is represented as a red star.

Regarding the heatmap, we can observe that this particular VALP found difficulties at generating samples of some classes, from which the 9-th is the worst case. The VALP was unable to generate any sample from this category according to MobileNet. The next worst case is class 2, with 399 generations, and the rest have at least 511 representatives out of 10.000 generations (A perfect model would have produced 1.000 from each category).

\begin{figure}
\begin{center}
\resizebox {8.8cm} {!}{\includegraphics{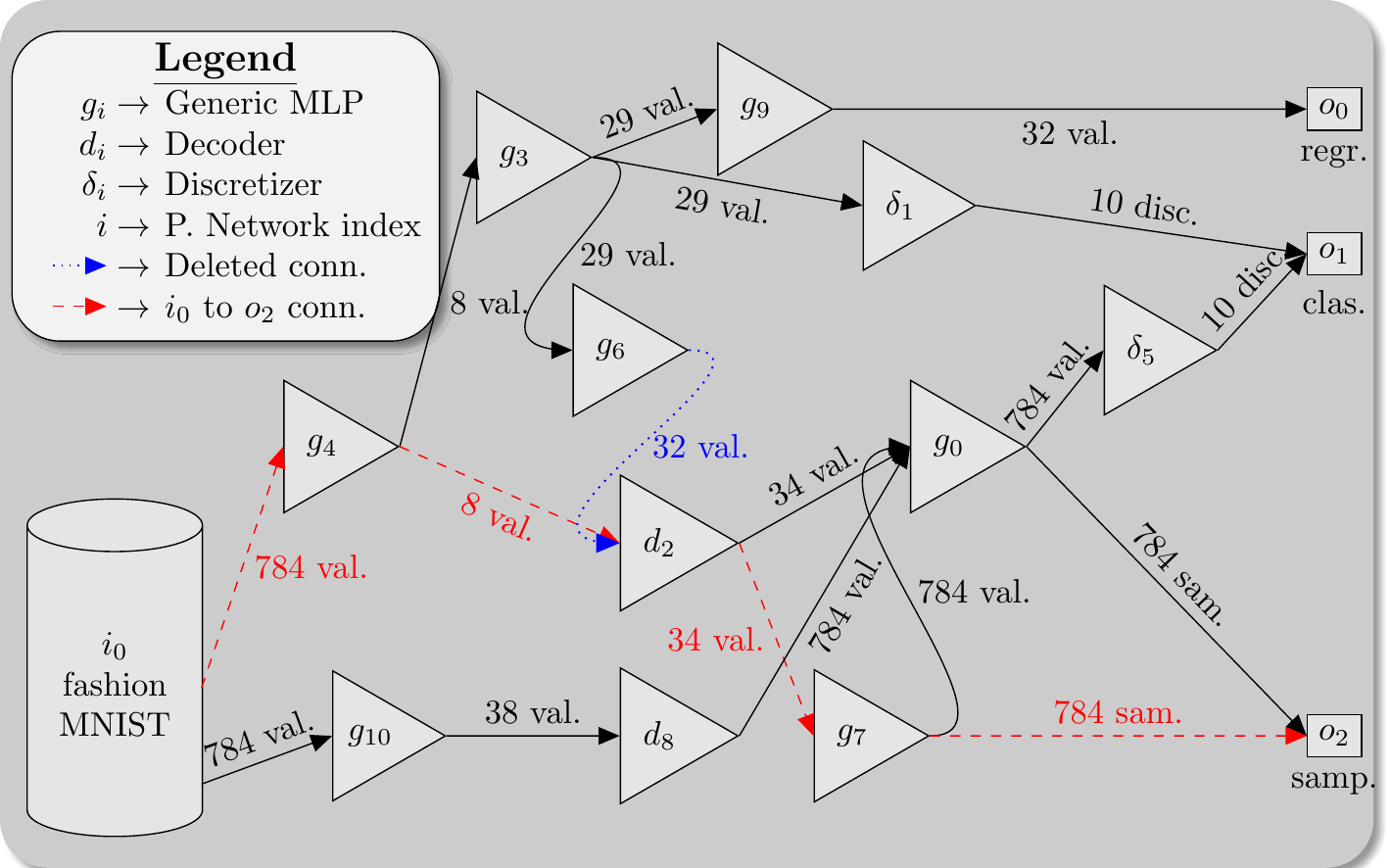}}
\caption{Schematic representation of the VALP configuration that showed the maximum conditioning from the input example in the produced sample. The arrow tags represent the size of the data; how many variables there are, ``val." stands for numeric values, ``disc." are discrete values, and ``sam." stands for samples. The dashed red lines denote the conditioning path. The dotted blue lines denote connections that are removed and replaced with $\mathcal{N}(0,I)$ when running the model.}
\label{fig:topAcc}
\end{center}
\end{figure}

Fig.~\ref{fig:topAcc} contains a schematic representation of the configuration of the VALP represented with the red star in Fig.~\ref{fig:entracc}, which, it is worth noting, achieved $85\%$ and $0.00053$ in classification accuracy and in regression MSE, respectively.

In this figure, we can observe that there are two decoders in the model, $d_2$ and $d_8$. The samples produced by $d_8$ \textit{only} suffered one transformation before being used for $o_2$. The conditioning part happens with $d_2$, as it receives two inputs while training, and only one of them is deleted in the feed-forward phase (represented as a dotted blue line). This enables the VALP to recycle a piece of information introduced in $i_0$ using the path represented in dashed red arrows, to generate the samples, ultimately producing conditioned examples.

\begin{figure}
\begin{center}
\includegraphics[width=8.8cm,trim={3pt 1pt 2pt 3pt},clip]{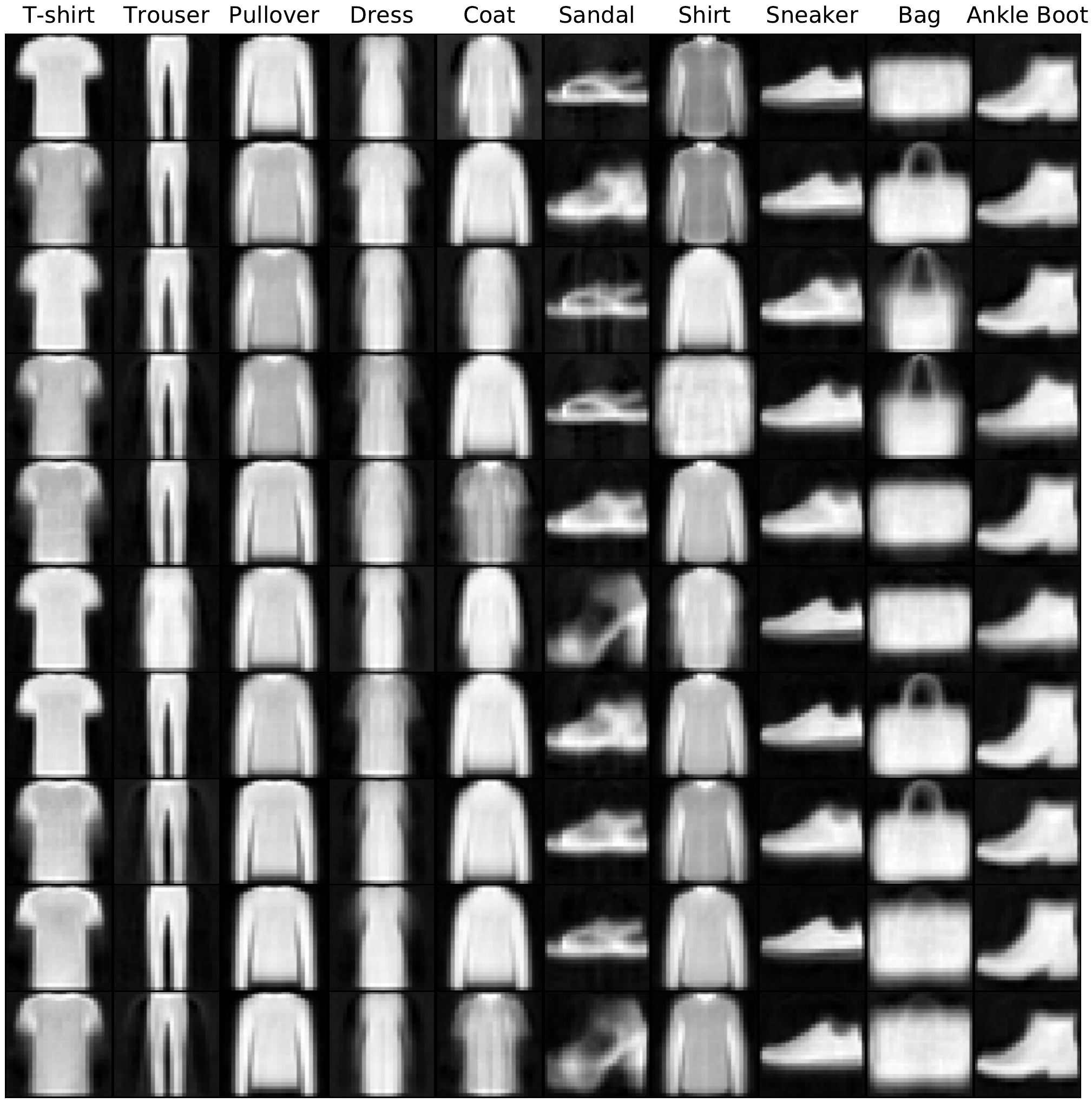}
\caption{Randomly chosen samples generated by the model represented with the red star in Fig.~\ref{fig:entracc}. The column names show the class the MobileNet assigned to them.}
\label{fig:imgs}
\end{center}
\end{figure}

Determining whether a model suffers from local mode collapsing is harder than the global type. In Fig.~\ref{fig:imgs}, we have displayed 10 random generations of the model represented with the red star in Fig.~\ref{fig:entracc} from each class they were conditioned towards.

From this figure, three different patterns can be identified. The first pattern consists of particularly weak samples. We can observe some blurry examples of Bags (specially the handles), Trousers (the sixth example), and Dresses (blurry sleeves). However, in most cases, it is possible to idenify what class each image belongs to.

Then, it is possible to appreciate a second pattern, which are those classes for which the generated samples are easily identifiable, even though they look very similar to each other. A good example of this is the Trouser class, which generated characteristic trouser images, but which are very similar to each other.

Finally, there is the third pattern, which are those classes that had identifiable generated samples, while keeping differences between them. Clear examples of this \textit{good} generation are the Sandal, Ankle Boot, and, in some cases, Bag classes.

Additionally, we can observe in this figure that, even though MobileNet failed to classify any generation as an Ankle Boot (classifying them as Sandal or Sneaker), the Ankle Boots were present among the generations of the model.

\section{Open Challenges}\label{sec:futurework}

After having empirically shown that the VALP is a viable approach to give a solution to the HMTL problem, the next step in the path is to identify the directions towards which the research over the VALP could be developed. In this section, we enumerate some of these research lines.

\subsection{Structure search}

In this paper a simple random procedure has been used to create VALP models. Every connection within the model in this experimentation is, both inter, and intra-network, random. An intelligent search over the space of possible structures would definitely help to improve the modelling capacity of the VALP. In this regard, evolutionary algorithms (EAs) arise as a promising strategy.

CoDeepNEAT \cite{miikkulainen_evolving_2017} is a recently introduced co-evolutionary algorithm that simultaneously evolves two types of model components: convolutional cells and empty structures. The former represents a way to extract high-abstraction features from data, the latter represents some flexible layout to place the cells. The algorithm outputs a model where the combination of the components is expected to be highly efficient for the task.  

Some similarities between the structures evolved by CoDeepNEAT and the VALP structures can be found (convolutional cells and primary networks, and empty structures and CALP connection structure), which suggests that the evolutionary algorithm could be used to optimize VALP structures. Even so, the application of this algorithm to the VALP would require some adjustments. The main difference resides in the fact that the convolutional cells do not have the \textit{identity} that characterizes the primary network. Therefore, CoDeepNEAT should be extended to simultaneously evolve VALP structures, and networks from each type separately, instead of considering all primary networks as \textit{equals}.

Another approach the VALP could benefit from is that based on path finding on supernetworks. The authors of \cite{fernando_pathnet:_2017} design a sizable DNN and collect sets of related, still homogenous, problems such as image recognition (MNIST \cite{lecun_gradient-based_1998}, CIFAR \cite{krizhevsky_learning_2009}, SVHN \cite{netzer_reading_2011}) and reinforcement learning (several Atari games). The EA consists of a population of agents that determine a \textit{path of neurons} across the randomly initialized DNN. Each individual is evaluated by training the parameters in the path of the network using the standard procedure -backpropagation- for the first task. Then, the next task is selected, and the weights of the connections that have been trained are frozen. Subsequently, the network is trained with the second task using another agent. This process is iteratively repeated until all tasks have been learned by the super network.

This algorithm could also be applied to the VALP. The model would be initialized using a vast amount of random primary networks and connections, always satisfying type restrictions. From this \textit{superVALP}, only a subset of primary networks would be trained to perform each of the tasks, i.e., the paths described by the agents would specify primary networks instead of neurons. This algorithm would need further adaptation, since it does not contemplate the quality of the elements in the path, and the primary neural networks usually have hyperparameters that need to be tuned to provide the maximum performance. This issue could be addressed using AutoML methods \cite{olson_evaluation_2016,kotthoff_auto-weka_2017,komer_hyperopt-sklearn:_2014}, that can optimize the structures of the primary networks. For example, when using an agent to train the superVALP to learn a task, the learning procedure could include an AutoML technique.

By extending and refining the data types used by VALP, expert knowledge can be introduced, in a way similar to the one used for extending other AutoML methods \cite{garciarena_evolving_2017}.

\subsection{New components}

In the VALP architectures considered in this work, all the primary networks were fully connected layers sequentially placed. Future VALP variants could implement other architectures (some of which have already been identified in this work) that offer an excelling performance in different areas. One clear example would be adding recurrence to the VALP for an improved addressing of problems with sequential data. One way to incorporate this concept to a VALP instance would be to allow recurrent connections within the primary networks, which was not contemplated in the random structure search used in our example. A more straightforward (and even complementary) way would consist of designing new primary networks that contained recurrent connections within themselves, e.g., primary networks consisting of LSTM \cite{gers_learning_1999} or GRU \cite{cho_learning_2014} cells, for example. 

CNNs would also play an important role in VALP, as they would allow the model to maximize its performance in problems with image (or similar) data. These type of networks have already been defined in this work as primary networks of a VALP. However, they are yet to be evaluated in this context.

\subsection{Loss functions}

We selected one single loss function per task type, namely, cross-entropy for classification, mean squared error for regression, and the log-likelihood for the sampling problem. However, more options exist for each of these problems, and these could be included in the optimization process of the model \cite{garciarena_evolved_2018,wang_evolutionary_2019}, since some loss functions could have positive contributions towards the optimal training of the model \cite{garciarena_evolved_2018} (for example, any of the GAN loss functions for the generative task \cite{goodfellow_nips_2016}). This incorporation would remove the necessity for a Generic MLP-Decoder structure in a sampling VALP. Not having a Decoder in a VALP configuration would mean not having to change the model structure (connection deletion) once it has been trained. This would be an upgrade in terms of flexibility.

Linked to the loss function selection topic, the weights applied to each of the sub-loss functions in the overall loss function have also been manually selected, after observing that the KL component could neutralize the effectiveness of the gradient descent algorithm. These parameter selection issues could be taken care of in the aforementioned optimization procedure, as a parameter tuning. Moreover, the multi-loss-function issue could also be addressed as a multi-objective problem of different VALP configurations that offer performances of varying quality over the different loss functions.

Finally, this work has expanded MTL to the combination of different types of tasks: prediction (regression and classification), and data generation. Other popular paradigms, such as reinforcement learning, have not been included. Combining so many loss functions of different natures and training them all in the same model presents itself as a very challenging task.

\section{Conclusions}\label{sec:conclusions}

In this paper, we have addressed the heterogeneous variant of the multi-task learning problem; the HMTL. This problem consists of training a single model to perform several tasks at the same time, these tasks being of different natures (e.g., regression and data sampling). To deal with this problem, we have proposed the innovative VALP model, a DNN-based approach. We have firstly provided a formal definition of the approach to lay the groundwork over which several different work directions can be developed. The main strength of the VALP is its capacity to manage different kinds of sub-DNNs and loss functions, which enables the model to produce different types of data that accurately approximate any distribution, using an optimization procedure over the different loss functions.

In this work, we have defined and focused on the fashion-MNIST HMTL problem, which consists of three different tasks (classification, regression, data generation). A particular VALP implementation has been designed to fit the particularities of said problem. A random search over the many-dimensional search space showed that the VALP can effectively and simultaneously carry out various tasks of different types, which also involve loss functions of completely different nature.

More specifically, we found VALP configurations that were able to optimize the classical prediction tasks (classification and regression), while still producing reasonably good results at data sampling. Some configurations were even able to partially avoid one of the most concerning issues in the generative community, mode collapsing. It is remarkable that VALP models whose architectures were randomly designed were able to obtain such strong results in both classification and regression, while still having reasonably good results in data generation.

Finally, we have enumerated a collection of detailed future research lines that the newly created VALP model can benefit from. Applying techniques that have produced high-quality results in other models to the VALP will help to determine where the strengths and limitations of the model lie.

\ifCLASSOPTIONcaptionsoff
  \newpage
\fi

\bibliographystyle{abbrv}
\bibliography{Zotero}

\begin{IEEEbiographynophoto}{Unai Garciarena}
received his bachelor degree in computer science in 2015, before obtaining his masters degree in computer engineering and intelligent systems, in 2016, both in the University of the Basque Country (UPV/EHU). He enrolled his Ph.D. studies in 2017 with the University of the Basque Country (UPV/EHU). His principal research interests are generative modeling, supervised classification, and optimization.
\end{IEEEbiographynophoto}

\begin{IEEEbiographynophoto}{Alexander Mendiburu}
received the Ph.D. degree from the University of the Basque Country, Spain, in 2006. Since 1999, he has been a Lecturer with the Department of Computer Architecture and Technology, University of the Basque Country. His current research interests include evolutionary computation, probabilistic graphical models, and parallel computing.
\end{IEEEbiographynophoto}

\begin{IEEEbiographynophoto}{Roberto Santana}

received the B.S. degree in computer science and the Ph.D. degree in mathematics from the University of Havana,  Havana, Cuba, in 1996 and 2005, respectively, and the Ph.D. degree in computer science from the University of the Basque Country, San Sebastian–Donostia, Spain, in 2006. He is a  Researcher  with the University of the Basque Country. His research interests include machine learning, evolutionary computation, probabilistic graphical models, and neuroscience.
\end{IEEEbiographynophoto}

\end{document}